\documentclass[11pt]{article}

\usepackage[preprint]{acl}
\usepackage{makecell}
\usepackage{times}
\usepackage{latexsym}
\usepackage{booktabs}
\usepackage[T1]{fontenc}

\usepackage[utf8]{inputenc}
\usepackage{CJKutf8}

\usepackage{xcolor}
\usepackage{tcolorbox}
\usepackage{ulem}
\usepackage{graphicx}

\usepackage{microtype}
\usepackage{multirow}
\usepackage[table]{xcolor}
\usepackage{array}
\usepackage{inconsolata}

\usepackage{graphicx}
\usepackage{amsmath}
\usepackage{algpseudocode}
\usepackage{amssymb}
\usepackage{pifont}     
\newcommand{\ours}{\textproc{\textbf{Mars}}}

\definecolor{boxborder}{RGB}{0, 0, 0}
\definecolor{headergray}{RGB}{235, 235, 235}
\definecolor{hlbluecolor}{RGB}{222, 235, 247}
\definecolor{hlpurplecolor}{RGB}{230, 225, 245}
\definecolor{arrowcolor}{RGB}{135, 175, 215}
\definecolor{orangetext}{RGB}{220, 150, 60}
\definecolor{hlredcolor}{RGB}{250, 215, 215}
\definecolor{hlgreencolor}{RGB}{215, 240, 215}
\definecolor{tagblue}{RGB}{60, 110, 180}
\definecolor{tagpurple}{RGB}{120, 80, 160}
\definecolor{taggreen}{RGB}{70, 140, 90}
\definecolor{tagorange}{RGB}{210, 120, 50}

\newcommand{\hlblue}[1]{%
    \bgroup
    \markoverwith{\textcolor{hlbluecolor}{\rule[-0.6ex]{2pt}{2.5ex}}}%
    \ULon{#1}%
}

\newcommand{\hlpurple}[1]{%
    \bgroup
    \markoverwith{\textcolor{hlpurplecolor}{\rule[-0.6ex]{2pt}{2.5ex}}}%
    \ULon{#1}%
}

\newcommand{\hlred}[1]{%
    \bgroup
    \markoverwith{\textcolor{hlredcolor}{\rule[-0.6ex]{2pt}{2.6ex}}}%
    \ULon{#1}%
}

\newcommand{\hlgreen}[1]{%
    \bgroup
    \markoverwith{\textcolor{hlgreencolor}{\rule[-0.6ex]{2pt}{2.6ex}}}%
    \ULon{#1}%
}

\title{\ours: Training Multilingual Search Agents for Retriever-Free Evidence Acquisition}

\author{
    \textbf{Rui Qi}\textsuperscript{1,2},
    \textbf{Yunlong Liang}\textsuperscript{3},
    \textbf{Yufeng Chen}\textsuperscript{1,2}\footnotemark[2], 
    \textbf{Jinan Xu}\textsuperscript{1,2},
    \textbf{Fandong Meng}\textsuperscript{3},
    \textbf{Kaiyu Huang}\textsuperscript{1,2}\footnotemark[2] \\
    \\
    \textsuperscript{1}Key Laboratory of Big Data \& Artificial Intelligence in Transportation \\
 (Beijing Jiaotong University), Ministry of Education \\
    \textsuperscript{2}School of Computer Science and Technology, Beijing Jiaotong University \\
    \textsuperscript{3}Pattern Recognition Center, WeChat AI, Tencent Inc., China \\
    \texttt{\{cherry,kyhuang\}@bjtu.edu.cn}
\\
}

\begin{document}
\maketitle

\renewcommand{\thefootnote}{\fnsymbol{footnote}}
\footnotetext[2]{Corresponding author.}
\begin{abstract}
Existing multilingual retrieval-augmented generation relies heavily on learned retrievers and similarity-based ranking from multilingual collections, which are vulnerable to cross-lingual retrieval errors and indexing overhead.
In this paper, we explore retriever-free multilingual search, where the model directly interacts with raw corpora through executable shell commands.
This paradigm shifts the bottleneck to multilingual search action generation, where exact search is sensitive to entity surface forms and outcome-only supervision provides limited guidance for multi-turn search.
To address these challenges, we propose \textbf{\ours}, a multilingual search agent in direct corpus interaction~(DCI) with reinforcement learning~(RL).
\ours~introduces \textbf{corpus-native entity grounding} to recover reliable lexical anchors from corpus evidence, mitigating failure to retrieve evidence.
Meanwhile, we present a \textbf{redundancy-aware search reward} to encourage concise and effective search trajectories, addressing training collapse in RL of multilingual scenarios.
Experiments show that \ours~consistently outperforms strong retrieval-based and DCI baselines, while demonstrating robust cross-lingual generalization and substantially reducing redundant corpus interactions. Our code is available at \url{https://anonymous.4open.science/r/MARS-0382/}
\end{abstract}

\section{Introduction}

Multilingual search agents have demonstrated considerable promise in addressing complex information needs~\cite{huang2025surveylargelanguagemodels,wang-etal-2025-x}.
By extending the conventional ``one-shot'' retrieval-augmented generation (RAG) pipeline into an iterative evidence acquisition process, these agents enable models to interact with retrieval environments and dynamically refine their search strategies based on intermediate observations~\cite{jin2025searchr1trainingllmsreason, li-etal-2025-search}.
This agentic paradigm has been increasingly adopted in multilingual RAG, where its effectiveness ultimately hinges on accurately retrieving the evidence required to establish the target fact from large-scale, cross-lingual collections~\cite{qi-etal-2026-language, lee-etal-2026-coral}.
\begin{figure}[!t]
    \centering
    \includegraphics[width=0.95\linewidth]{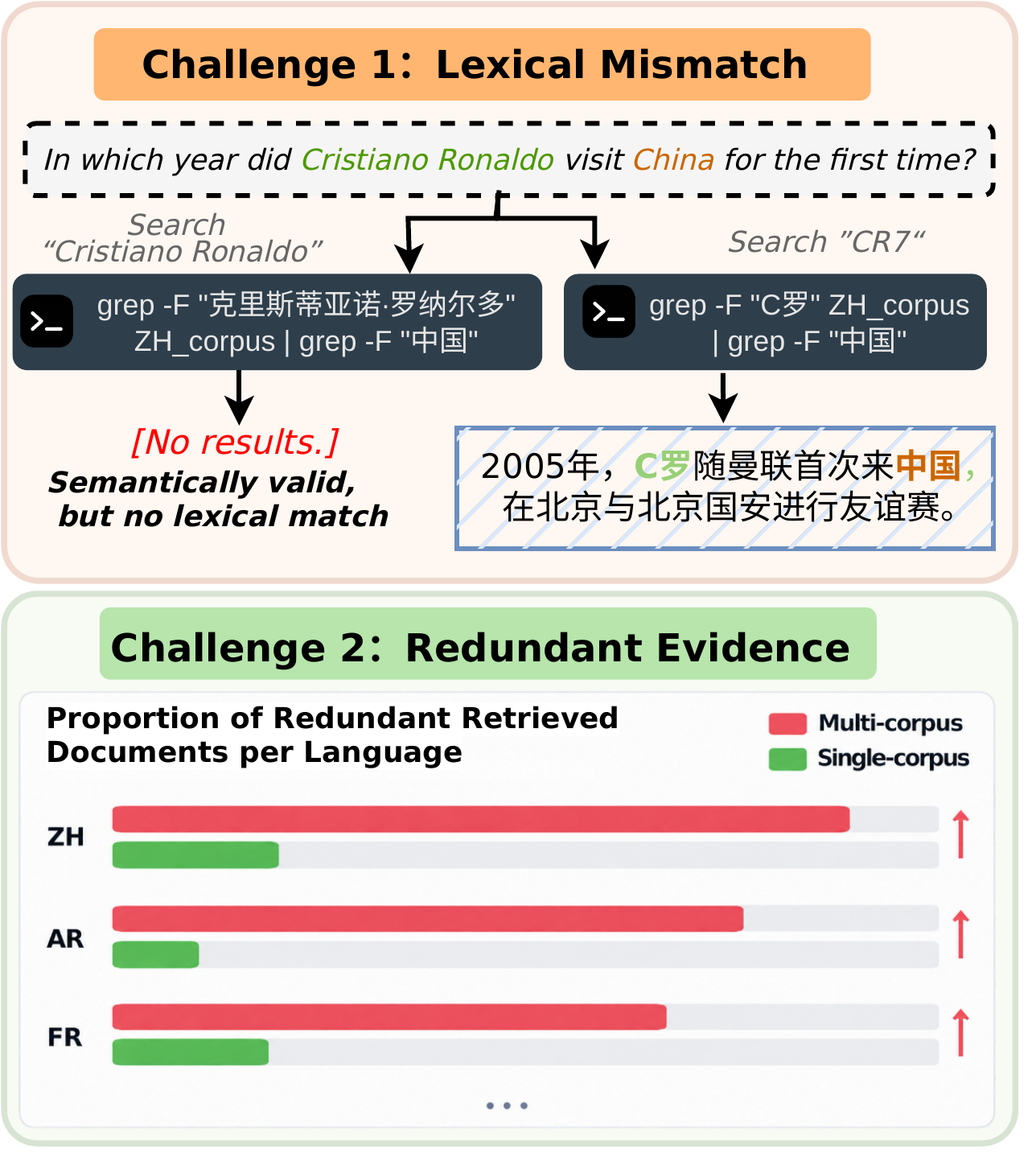}
    \caption{Illustration of lexical mismatch and trajectory optimization challenges in multilingual DCI.
Challenge 2 compares multi-corpus and single-corpus search, where the non-redundant ratio denotes the fraction of retrieved documents that are unique within a trajectory for each language.
}
    \label{fig:head}
\end{figure}

These approaches rely on research in developing retrieval models, such as semantic scoring based on dense or sparse representations~\cite{thakur2021beirheterogenousbenchmarkzeroshot,zhang-etal-2023-miracl}, and retrieval performance remains poor for low-resource languages, limiting the effectiveness of multilingual search agents.
More importantly, large-scale multilingual collections further exacerbate the limitations of similarity-based retrieval, particularly the problem of retrieving documents that are similar in representation space but irrelevant to the underlying information need~\cite{zhang-etal-2025-utility}.
Systematic shifts in embedding distributions across languages can cause a document with a lower score to be more relevant than a higher-scoring document written in another language.
In addition, embeddings and indexing for collections incur substantial costs in terms of storage, construction, and continuous updates. 
As a result, these limitations underscore the need to explore new paradigms for information acquisition in multilingual settings.


Executable search tools provide a direct interface to large-scale corpora, enabling code agents to locate information through explicit operations~\cite{NEURIPS2024_5a7c9475}. 
Inspired by this, this paper explores a retriever-free multilingual search in which the model interacts directly with raw corpora through shell commands. 
This interface enables precise lexical matching and supports compositional constraints through executable search pipelines, providing the model with finer-grained control over evidence acquisition than similarity-based retrieval.
Direct corpus interaction~(DCI)~\cite{li2026semanticsimilarityrethinkingretrieval} transfers the bottleneck from relevance scoring to the formulation of precise multilingual search actions, which remains a huge challenge as illustrated in Figure~\ref{fig:head}. 
First, exact search is sensitive to lexical forms, while entity names often vary across languages due to translation, transliteration, and aliases. DCI can frequently miss relevant evidence when the query is expressed in a language different from that of the corpus. 
Second, the model needs to learn an effective cross-lingual search strategy that determines when to reformulate queries, switch to a corpus in another language, refine intermediate results, or terminate the search process.

To address these challenges, we introduce~\ours, the first DCI framework for \textbf{m}ultilingual \textbf{a}gentic \textbf{r}etreiever-free \textbf{s}earch, which enables direct interaction with raw multilingual corpora through executable search actions rather than relying on learned retrievers and cross-lingual similarity scoring.
In particular, we propose a Utility-Aware Retrieval Policy Optimization~(URPO) method based on GRPO~\cite{shao2024deepseekmathpushinglimitsmathematical}. 
This method introduces \textbf{corpus-native entity grounding} and \textbf{redundancy-aware search rewards} for URPO, achieving stable optimization during command refinement and coordinating the learning of when agents should adaptively switch across multilingual corpora.

Specifically, corpus-native entity grounding extracts language-specific titles and aliases from retrieved evidence and reuses them as reliable search keys, reducing exact-match failures caused by surface-form mismatches. 
The redundancy-aware search reward uses search conciseness as a lightweight signal to enable the proposed framework to autonomously determine language-specific corpus search actions based on intermediate observations.
Experimental results show that our method consistently enhances the abilities of backbones for multilingual RAG and robust performance in various experimental settings compared to baselines, while reducing ineffective and redundant corpus interactions. Our contributions are as follows:
\begin{itemize}
    \item We propose ~\ours, the first DCI framework for multilingual RAG, which enables LLMs to acquire evidence directly from multilingual corpora without relying on learned retrievers and achieves substantial improvements over strong MRAG baselines.
    \item Our proposed Utility-aware Retrieval Policy Optimization mitigates surface-form mismatches and regulates multilingual search trajectories, leading to more reliable corpus interaction as language diversity increases.
    \item Experiments show that our method transfers effectively to both seen and unseen languages and remains robust under sparse EM supervision, where existing multilingual agentic RAG baselines exhibit severe training instability and performance collapse.
\end{itemize}

\begin{figure*}
    \centering
    \includegraphics[width=0.95\linewidth]{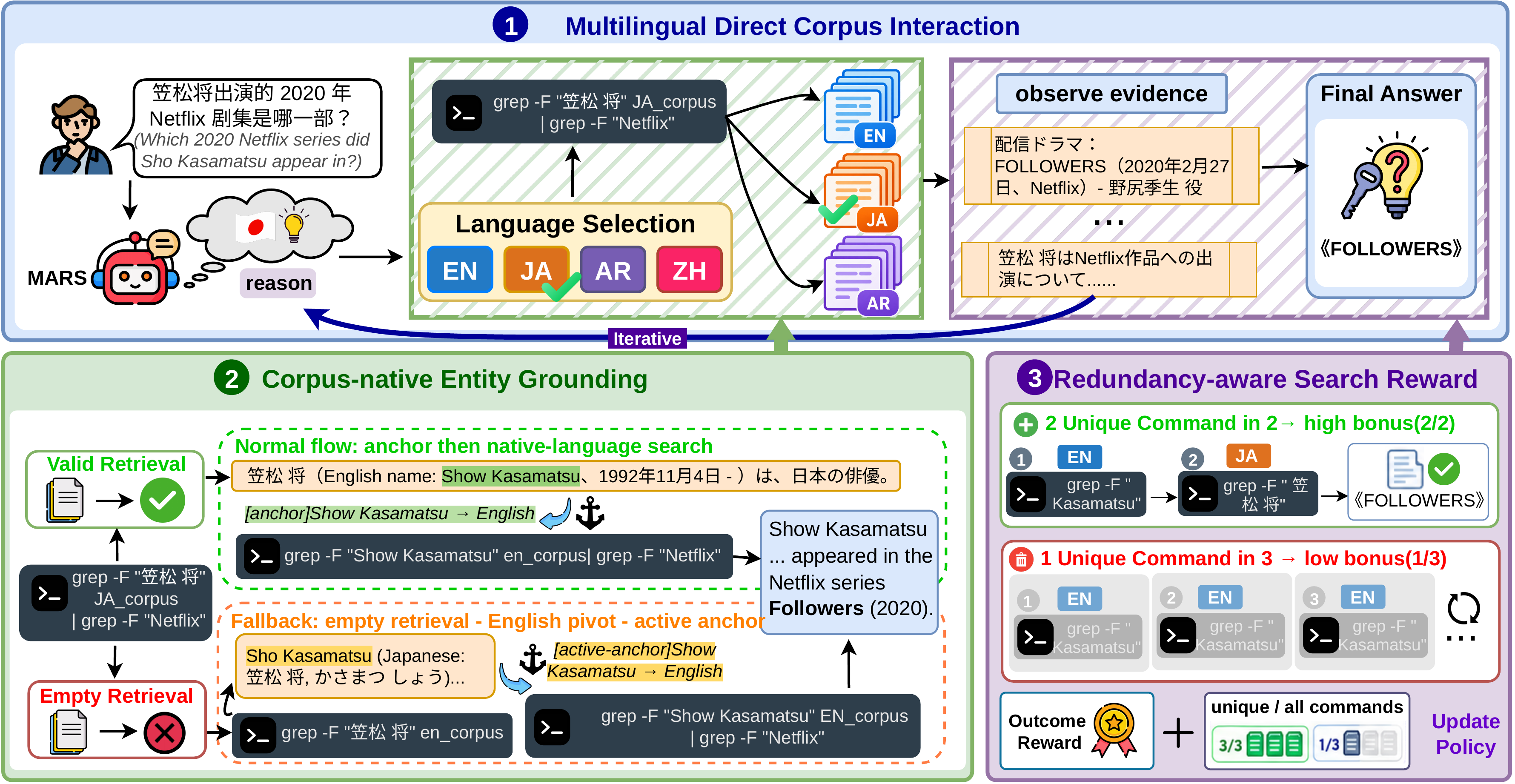}
    \caption{Illustration of \ours~for multilingual DCI.
The policy directly interacts with language-specific corpora in URPO, while corpus-native grounding supplies reusable lexical anchors and redundancy-aware search encourages successful, non-redundant search trajectories.
}
    \label{fig:framework}
\end{figure*}

\section{Method}
\subsection{Multilingual Direct Corpus Interaction}
As shown in Figure~\ref{fig:framework}, our rollout adopts an iterative search process that allows the model to directly interact with multiple language-specific corpora and incorporate language selection into the search decisions, rather than a pre-defined retrieval schedule.
Given a multilingual query $q$ and a set of corpora $\mathcal{C}=\{C^{\ell}\}_{\ell\in\mathcal{L}}$, where $\mathcal{L}$ denotes the available languages, the model performs multi-turn reasoning and corpus interaction until it produces a final answer or reaches the maximum interaction budget.
At each turn $i$, the policy first generates a reasoning trace $t_i$ within the \texttt{<think>} block based on the query and previous interaction history.
When external evidence is required, the model selects a target language $\ell_i$ and generates an executable shell command $c_i$ within \texttt{<tool\_call>} to search the corresponding corpus $C^{\ell_i}$.
Different from retriever-based MRAG, where a generated query is passed to a retriever that returns a ranked list of passages, our model directly specifies how the raw corpus should be searched through executable operations.
Formally, we define a multilingual search action as
\begin{equation}
    a_i = (\ell_i, c_i),
\end{equation}
where $\ell_i \in \mathcal{L}$ specifies the target corpus and $c_i$ denotes the executable search command serialized in the
\texttt{<tool\_call>} block.
Conditioned on the query and preceding trajectory $\tau_{<i}$, the model generates
\begin{equation}
    (t_i, a_i) \sim
    \pi_{\theta}(\cdot \mid q, \tau_{<i}).
\end{equation}
The execution engine then applies $c_i$ directly to $C^{\ell_i}$ and returns the resulting corpus observation
\begin{equation}
    o_i = \mathcal{E}(c_i, C^{\ell_i}),
\end{equation}
which is provided to the model through \texttt{<tool\_response>} and appended to the interaction history.
The complete search trajectory is therefore represented as
\begin{equation}
    \tau = \{(t_i,a_i,o_i)\}_{i=1}^{T},
\end{equation}
where $T$ is bounded by the maximum number of interaction turns.
Based on the accumulated observations, the model can reformulate its search, switch to another language corpus, further refine the retrieved evidence, or terminate the interaction by generating the final answer $\hat{y}$ within \texttt{<answer>}.


\subsection{Corpus-Native Entity Grounding}
\label{sec:grounding}
Direct corpus interaction relies on explicit lexical matching, making retrieval highly sensitive to entity surface forms.
When searching a corpus in a language different from that of the query, the model should formulate the search command using a corresponding target-language entity form.
However, translated or reformulated expressions may differ from the canonical form used in the target corpus due to transliteration, aliases, or orthographic conventions.
Thus, even a semantically correct translation may fail to match the entity's corpus-native form.

In some cases, entity-bearing passages expose a corpus-native entity form together with alternative names, such as Latin-script or English aliases.
These provide language-specific search keys for the corresponding collections.
Based on this observation, we introduce \textbf{corpus-native entity grounding}, which derives reusable lexical anchors directly from corpus evidence.
The lexical anchor is categorized into two types based on the current evidence acquisition status: (1) Grounding from corpus evidence and (2) Grounding under empty retrieval.

\noindent \textbf{Grounding from Corpus Evidence.}
Given a search action at turn $i$, let $D_i=\{d_j\}_{j=1}^{K}$ denotes the returned corpus candidates, where each candidate $d_j=(\mathrm{id}_j,\mathrm{text}_j)$ contains document id and textual content.
To avoid extracting entity forms from passages that merely mention the queried entity, we estimate a grounding confidence score $\boldsymbol{\phi}(d)$, 
where \(\boldsymbol{\phi}(d)\) calculates the possibility of the passage is likely to explicitly introduce the entity, such as lead-position patterns, entity-introduction structures, and alternative entity forms. The candidated passage is defined as:
\begin{equation}
    d_i^{*}=\arg\max_{d\in D_i}\boldsymbol{\phi}(d), \boldsymbol{\phi}(d_i^{*})\geq\gamma
\end{equation}

From the selected passage, we extract a set of explicitly observed entity forms
\begin{equation}
    g(d_i^{*})={G}_i=\{e_{i,m}\}_{m=1}^{M_i},
\end{equation}
where $e_{i,m}$ denotes the span that is taken into account for calculating the score $\boldsymbol{\phi}(d)$.

\noindent \textbf{Grounding under Empty Retrieval.}
The above procedure requires at least one useful corpus observation.
In the situation of empty retrieval, we apply the same grounding operator to a high-resource pivot corpus $C^{\ell_p}$ using the search terms extracted from the failed command.
In our experiments, we use English as the pivot language due to its broad corpus coverage.

Finally, the resulting signal is appended to the original tool observation: $\tilde{o}_i=o_i\oplus G_i,$
where $\oplus$ denotes observation augmentation.
The grounding signal is exposed through an explicit \texttt{[anchor/active-anchor]} tag. 
All grounding signals are derived directly from corpus observations without external translators, entity linkers, dictionaries, or additional language models.

\noindent \textbf{Grounding Utilization Reward.}
The availability of grounding signals does not guarantee that they will be leveraged by the policy in subsequent search, so we introduce a lightweight process reward for successful reuse of corpus-derived entity forms.
For a trajectory $\tau$, the grounding-utilization reward is
\begin{equation}
    R_{\mathrm{gro}}(\tau)
    =
    \lambda_g
    \cdot
    \mathbb{I}[\mathrm{EM}(\hat{y},y)=1\land  e_{i,m}\in \tau].
\end{equation}
It ensures that the bonus is assigned only when grounding reuse contributes to a correct trajectory, encouraging the behavior chain of acquiring, reusing, and successfully exploiting corpus-native entity forms.

\subsection{Redundancy-Aware Search Reward}
Corpus-native grounding improves the accessibility of multilingual evidence, but obtaining useful evidence still depends on the iteratively generation of effective search commands based on agents, where DCI delegates the retrieval process to model-generated executable commands.
Previous direct search training has primarily relied on outcome-based rewards to evaluate final answers, providing limited supervision on how search commands evolve along a trajectory in multilingual scenarios that require frequent switching across language-specific corpora and diverse evidence acquisition.
Without explicit guidance on command transitions, the policy has difficulty learning when and how to refine its search, and can instead fall back on repeated or near-duplicate commands that avoid the uncertainty of introducing new search terms and potentially noisy, resulting in ineffective evidence acquisition.
We therefore introduce a \textbf{redundancy-aware search reward}, which uses search conciseness as a lightweight signal to suppress inefficient evidence acquisition while preserving the flexibility required for legitimate cross-lingual exploration.

\noindent \textbf{Redundancy-Aware Search Modeling.}
For a trajectory $\tau$, let $\mathcal{C}_{\tau}=\{c_1,\ldots,c_n\}$ denote its sequence of executable search commands.
The similarity between two commands is measured by Jaccard similarity:
\begin{equation}
    \operatorname{sim}(c_i,c_j)
    =
    \frac{|B(c_i)\cap B(c_j)|}
    {|B(c_i)\cup B(c_j)|}.
\end{equation}
Two commands are treated as near-duplicates when $\operatorname{sim}(c_i,c_j)>\eta$, then we merge them into a single one.
Let $n_d$ denote the number of distinct commands after near-duplicate merging and $n$ is the total number.
The redundancy-free ratio is defined as
\begin{equation}
    \rho(\tau)=\frac{n_d}{n},
\end{equation}
where a lower value indicates that a larger fraction of the trajectory is spent revisiting similar search.
The URPO rewards non-redundant search only when the trajectory successfully answers the question:
\begin{equation}
    R_{\mathrm{red}}(\tau)
    =
    \lambda_r
    \cdot
    \mathbb{I}[\mathrm{EM}(\hat{y},y)=1]
    \cdot
    \rho(\tau).
\end{equation}
It prevents unsuccessful trajectories from receiving positive supervision merely by generating diverse commands.
Meanwhile, $\rho(\tau)$ discourages repeated search patterns without explicitly constraining which language corpus or search operation the policy should choose.
This allows the model to retain necessary cross-lingual exploration while favoring successful trajectories with less redundant evidence acquisition.

\subsection{Utility-Aware Retrieval Policy Optimization}
We combine the redundancy-aware reward with the grounding-utilization reward in Section~\ref{sec:grounding} and the base outcome reward for optimization:
\begin{equation}
    R(\tau)
    =
    \underbrace{\mathrm{EM}(\hat{y},y)-p(L_{\tau})}_{R_{\mathrm{base}}}
    +
    R_{\mathrm{gro}}(\tau)
    +
    R_{\mathrm{red}}(\tau),
\end{equation}
where $p(L_{\tau})$ denotes the trajectory-length penalty.
The length penalty controls excessive interaction, while $R_{\mathrm{red}}$ specifically regularizes repeated command-level search behavior.
The resulting trajectory reward is used for GRPO optimization, encouraging the policy to acquire useful multilingual evidence through more concise and informative command transitions.

\section{Experiments}
\subsection{Experimental Setup}
\noindent \textbf{Datasets.} We evaluate the performance of methods on two multilingual RAG benchmarks, \textbf{MKQA}~\citep{longpre-etal-2021-mkqa} and \textbf{MLQA}~\citep{lewis-etal-2020-mlqa} as they best represent multilingual open-ended question-answering tasks.
\textbf{MKQA} is a multilingual open-domain QA benchmark constructed from 10K questions in NQ dataset and provides parallel queries in multiple languages.
Following prior multilingual RAG settings, we use 13 languages and divide them into seen and unseen groups to evaluate cross-lingual transfer.
\textbf{MLQA} is a multilingual extractive QA benchmark derived from SQuAD and also provides parallel instances in seven languages.
The statistic of two benchmarks is shown in Table~\ref{tab:language_split} and we use two benchmarks to further evaluate the generalization abilities of the learned corpus interaction policy to various languages and data.

\begin{table}[t]
\centering
\small
\begin{tabular}{llll}
\toprule
Dataset & Split & \# / Lang. & Lang. \\
\midrule
MKQA & Train & 1000 &
\makecell[l]{En, Ar, Fi, Fr, \\Ja, It, Ru, Zh }\\
\cmidrule{2-4}
     & Test & 1277 &
\makecell[l]{Train Lang. + De, Ko,\\ Es, Pt, Th (Unseen)} \\
\midrule
MLQA & Train & 1000 &
\makecell[l]{En, De, Es, Hi,\\ Vi, Zh, Ar}
 \\
\cmidrule{2-4}
 & Test & 1000 & Train Lang.\\
\bottomrule
\end{tabular}
\caption{Language coverage for seen and unseen categorization.}
\label{tab:language_split}
\end{table}

\noindent \textbf{Baselines.} 
We compare \ours~with several representative baselines from both retrieval-based RAG and direct corpus interaction.
For retrieval-based methods, we include (1) Naive \textbf{RAG} and (2) \textbf{D-RAG}~\citep{ranaldi-etal-2025-improving-multilingual} as representative static RAG approaches, and (3) \textbf{LcRL}~\citep{qi-etal-2026-language} as a multilingual agentic RAG method.
For DCI, we include (1) Naive \textbf{DCI}~\cite{li2026semanticsimilarityrethinkingretrieval} as a training-free baseline and (2) \textbf{Grepseek}~\citep{salemi2026grepseektrainingsearchagents} as a representative post-trained DCI method.
We additionally report \textbf{Direct} inference without external knowledge as a reference.
Implementation details for all baselines are provided in Appendix~\ref{app:baselines}.

\noindent \textbf{Evaluation Metrics.}
We evaluate answer qualities using \textbf{exact-match (EM)} that measures whether the predicted answer exactly matches the reference answer after standard normalization, and we also follow the prior study~\cite{ranaldi-etal-2025-improving-multilingual,ranaldi-etal-2026-multilingual} to evaluate \textbf{flexible exact-match~(fEM)} accuracy, which is based on whether or not ground-truth answers are included in the generated answers.

\subsection{Main Results}

\begin{table}[!t]
    \centering
    \small
    \definecolor{colorMethods}{RGB}{255,255,255}
    \definecolor{colorSeen}{RGB}{226,243,246}
    \definecolor{colorUnseen}{RGB}{238,228,241}
    \resizebox{\columnwidth}{!}{
    \begin{tabular}{
        >{\columncolor{colorMethods}}l
        *{2}{>{\columncolor{colorSeen}}c}
        *{2}{>{\columncolor{colorUnseen}}c}
        *{2}{>{\columncolor{colorSeen}}c}
    }
        \toprule
        \multirow{2}{*}{\textbf{Methods}}
        & \multicolumn{2}{c}{\textbf{MKQA-Seen}}
        
        & \multicolumn{2}{c}{\textbf{MKQA-Unseen}}
         
        & \multicolumn{2}{c}{\textbf{MLQA}} 
         \\
        
         
        \cmidrule(lr){2-3}
        \cmidrule(lr){4-5}
        \cmidrule(lr){6-7}
        & \textbf{EM}
        & \textbf{fEM}
        & \textbf{EM}
        & \textbf{fEM}
        & \textbf{EM}
        & \textbf{fEM}\\
        \midrule

        \multicolumn{7}{c}{
            \textbf{\texttt{Qwen3.5-4B}}
        } \\
        \midrule

        Direct
        & 13.9 & 17.3 & 13.7 & 17.0 & 5.3 & 6.1 \\

        RAG
        & 24.0 & 30.8 & 21.9 & 27.6 & 10.4 & 12.7 \\

        D-RAG
        & 27.1 & 33.3 & 23.3 & 29.1 & 11.6 & 14.3 \\

        LcRL
        & 24.0& 28.9 & 21.0 & 27.1 & 7.6 & 10.1 \\

        DCI
        & 24.0 & 38.0 & 26.3 & 39.2 & 10.0 & 17.4 \\

        Grepseek
        & 33.9 & 47.5 & 30.5 & 45.5 & 13.9 & 31.7 \\

        \ours~
        & \textbf{36.3}$\dagger$ & \textbf{51.2}$\dagger$ & \textbf{33.2}$\dagger$ & \textbf{49.1}$\dagger$ & \textbf{16.0}$\dagger$ & \textbf{33.6}$\dagger$ \\

        \midrule

        \multicolumn{7}{c}{
            \textbf{\texttt{Qwen3.5-9B}}
        } \\
        \midrule

               Direct
        & 18.1 & 23.6 & 18.2 & 23.3 & 6.1 & 7.5 \\

        RAG
        & 25.4 & 33.3 & 23.9 & 31.5 & 11.1 & 14.1 \\

        D-RAG
        & 27.1 & 33.0 & 21.9 & 28.0 & 12.7 & 15.3 \\

        LcRL
        & 27.2 & 32.4 & 25.6 & 30.7 & 10.1 & 13.2 \\

        DCI
        & 26.8 & 46.3 & 29.0 & 48.0 & 11.2 & 21.4 \\

        Grepseek
        & 37.1 & 53.2 & 35.6 & 52.5 & 15.0 & 33.1 \\

        \ours~
        & \textbf{40.6}$\dagger$ & \textbf{55.5}$\dagger$ & \textbf{37.6}$\dagger$ & \textbf{54.1}$\dagger$ & \textbf{16.5}$\dagger$ & \textbf{34.9}$\dagger$ \\

        \bottomrule
    \end{tabular}
    }
    \caption{
    Main results (\%) of on MKQA and MLQA. We report the average scores of categorized languages, including \colorbox{colorSeen}{seen languages} of MKQA and MLQA, and \colorbox{colorUnseen}{unseen languages} of MKQA.
    The highest score under each base model is highlighted in \textbf{bold}, while $\dagger$ denote statistically significant improvement, compared to the best baseline (p < 0.05). Detail results of each language are provided in Appendix~\ref{app:detail_results}.
    }
    \label{tab:main_results}
\end{table}

As shown in Table~\ref{tab:main_results}, \ours~achieves better performance in all evaluated settings compared to all baselines.
The framework can also improve multilingual corpus interaction and consistently surpasses both DCI baselines, and the benefits are not tied to a particular backbone.
It indicates that multilingual direct search requires the policy not only to generate executable commands but also to align entity forms with language-specific corpora and continually adjust search actions across different collections.
Moreover, DCI-based methods generally outperform retrieval-based approaches on different model scales and evaluation metrics.
This suggests that directly operating on raw multilingual corpora can provide more effective evidence access than relying on a fixed retrieval pipeline, especially when the required facts are difficult to capture through similarity-based ranking.

\begin{table}[!t]
\centering
\small
\resizebox{\columnwidth}{!}{
\begin{tabular}{ccll}
\toprule
\textbf{$R_{\mathrm{gro}}$} 
& \textbf{$R_{\mathrm{red}}$} 
& \textbf{EM}  
& \textbf{fEM} \\
\midrule

\ding{55}
& \ding{55}
& 37.07 
& 53.23 \\

\ding{51}
& \ding{55}
& 39.79$_{\textcolor{red}{+2.72}}$
& 53.97$_{\textcolor{red}{+0.74}}$ \\

\ding{55}
& \ding{51}
& 38.97$_{\textcolor{red}{+1.90}}$
& 54.50$_{\textcolor{red}{+1.27}}$ \\

\rowcolor{gray!15}
\ding{51}
& \ding{51}
& \textbf{40.64}$_{\textcolor{red}{+3.57}}$ 
& \textbf{55.54}$_{\textcolor{red}{+2.31}}$ \\
\bottomrule
\end{tabular}
}
\caption{
Ablation results on MKQA.
Subscripts denote absolute improvements over the multilingual DCI baseline.
}
\label{tab:ablation}
\end{table}
\begin{table}[!t]
\centering
\resizebox{\columnwidth}{!}{
\begin{tabular}{lcccc}
\toprule
\textbf{Search Key}
& \makecell{\textbf{Empty}\\\textbf{Hit} $\downarrow$}
& \makecell{\textbf{First}\\\textbf{Hit} $\downarrow$}
& \makecell{\textbf{Correct}\\\textbf{Hit} $\uparrow$}
& \makecell{\\\textbf{EM} $\uparrow$} \\
\midrule
In-Trans
& 32.2 & \textbf{1.25} & 54.1 & 37.1 \\

Ex-Trans
& 39.0 & 1.33 & 54.9 & 38.3 \\

\rowcolor{gray!15}
\textbf{Ours}
& \textbf{24.3} & 1.30 & \textbf{56.7} & \textbf{40.3} \\
\bottomrule
\end{tabular}}
\caption{
Effects of search-key sources by different strategies on multilingual evidence access. ``First Hit'' denotes the number of rounds required for the first hit.
}
\label{tab:grounding_analysis}
\end{table}

\subsection{Ablation Studies}

As shown in Table~\ref{tab:ablation}, we further investigate the effectiveness of the proposed components by progressively augmenting the multilingual DCI baseline.
The corpus-native entity grounding yields a clear improvement in both EM and fEM, indicating that corpus-derived entity forms provide more reliable access to multilingual evidence under exact lexical search.
Meanwhile, the performance of using the redundancy-aware reward alone also shows that search-trajectory regularization provides independent benefits even without grounding.
Furthermore, the redundancy-aware policy optimization provides further gains based on the grounding-enhanced model.
The improvement shows that making relevant evidence accessible is insufficient and the policy must learn to refine executable commands and avoid ineffective search transitions during multi-turn interaction.
Overall, \ours~can achieve the best performance, demonstrating that corpus-native grounding improves evidence localization, while redundancy-aware optimization regulates the subsequent search trajectory.

\begin{table}[!t]
\centering
\small
\resizebox{\columnwidth}{!}{
\begin{tabular}{lcc}
\toprule
\textbf{Metric} 
& \textbf{Ours} 
& \textbf{w/o. $R_{\mathrm{red}}$} \\
\midrule
Repeated Search Rate $\downarrow$
& \textbf{4.89\%}
& 22.33\% \\

\# Search Commands $\downarrow$
& \textbf{2.56}
& 5.51 \\

\# Interaction Turns $\downarrow$
& \textbf{1.36}
& 2.27 \\
\bottomrule
\end{tabular}
}
\caption{
Effects of redundancy-aware search on multilingual DCI trajectories. We report the average of all queries. \# denotes the count.
Lower values indicate more concise and less redundant corpus interaction.
}
\label{tab:rfd_analysis}
\end{table}

\section{Analysis and Discussion}

\subsection{Effects of Corpus-Native Entity Grounding on Evidence Acquisition}

As shown in Table~\ref{tab:grounding_analysis}, we further investigate the improvements brought by corpus-native entity grounding in accessing multilingual evidence through exact lexical search.
We compare three sources of search keys: the model's internal translation using Prompt (\textbf{In-Trans}), an external translation API as a diagnostic baseline (\textbf{Ex-Trans}), and corpus-derived titles and aliases provided by our grounding mechanism.
The results show that corpus-native entity grounding substantially improves evidence accessibility.
The empty-hit rate decreases from 32.2 to 24.3, while the correct-entity hit rate increases from 54.1 to 56.7.
In contrast, the external translation API increases the empty-hit rate to 39.0 despite a small improvement in correct-entity coverage.
This result suggests that a linguistically plausible translation does not necessarily provide the lexical form required by the target corpus.
Corpus-derived titles and aliases instead provide search keys that are better aligned with the expressions actually used in the collection.
This improvement is also reflected in downstream performance, where the EM increases from 37.1 to 40.3.
Overall, our proposed method improves both the accessibility and correctness of multilingual evidence under exact search.

\begin{table*}[!t]
\centering
\small
\resizebox{\textwidth}{!}{
\begin{tabular}{lcccccc}
\toprule
\textbf{Methods}
& \textbf{Offline Index}
& \textbf{$\uparrow$ EM (\%)}
& \makecell{\textbf{$\downarrow$ Search}\\\textbf{Commands (s)}}
& \makecell{\textbf{$\downarrow$ Search}\\\textbf{Turns (\#)}}
& \makecell{\textbf{$\downarrow$ Search}\\\textbf{Duplicate Rate (\%)}}
& \makecell{\textbf{$\downarrow$ Response Tokens}\\\textbf{(Latency Proxy)}} \\
\midrule

D-RAG
& High (mE5)
& -- 
& --
& --
& --
& 6,920 \\

LcRL
& High (mE5)
& 27.23 
& --
& 2.07
& 6.2
& 3,532 \\

Grepseek
& None
& 37.07
& 4.22
& 1.87
& 16.4
& 4,853 \\

\rowcolor{gray!15}
\textbf{\ours}
& None
& \textbf{40.64}
& \textbf{2.56}
& \textbf{1.36}
& \textbf{4.9}
& \textbf{3,447} \\

\midrule
\makecell{\textbf{($\Delta$) \ours}\\vs. \textbf{Grepseek}}
& --
& \textbf{+3.2 pt}
& \textbf{$\downarrow$39\%}
& \textbf{$\downarrow$27\%}
& \textbf{$\downarrow$70\%}
& \textbf{$\downarrow$29\%} \\

\bottomrule
\end{tabular}
}
\caption{
Efficiency comparison between dense multilingual RAG and direct corpus interaction.
Response tokens are reported as a proxy for generation latency.
}
\label{tab:efficiency}
\end{table*}

\subsection{Effects of Redundancy-Aware Search Reward}

As shown in Table~\ref{tab:rfd_analysis}, we further explore the fine-grained impact of the redundancy-aware reward on the policy’s search actions by comparing the full method with a variant without $R_{\mathrm{red}}$.
The results demonstrate an increase in the proportion of duplicate search commands, from 4.89\% to 22.33\%, after the removal of the redundancy-aware search.
This indicates that without explicit redundancy-aware guidance, the policy tends to revisit similar search operations even when previous interactions have already provided comparable observations.
The same trend is reflected in the overall interaction cost.
Without $R_{\mathrm{red}}$, the average number of search commands increases from 2.56 to 5.51, while the average number of interaction turns rises from 1.36 to 2.27.
In contrast, our full model reaches its decisions with substantially fewer corpus interactions.
It shows that redundancy-aware optimization specifically encourages the policy to avoid revisiting similar search states and to acquire evidence through more concise command sequences.

\begin{figure}
    \centering
    \includegraphics[width=1\linewidth]{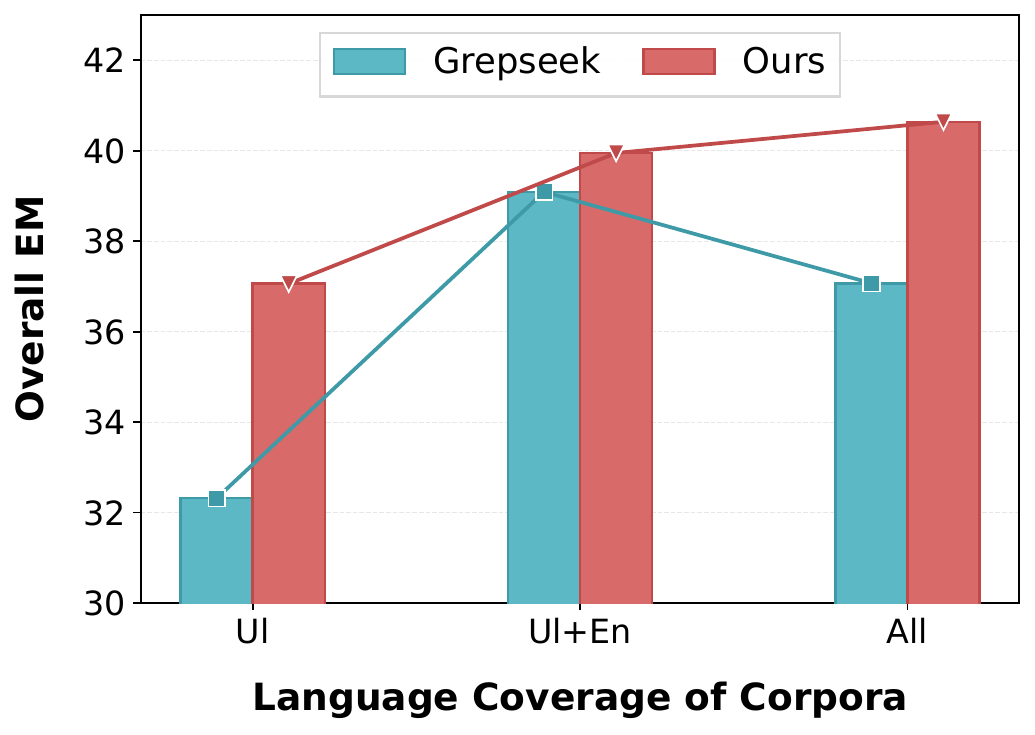}
    \caption{
Overall EM under increasing corpus-language coverage on MKQA.
\textit{Ul} denotes the corpus in the user-query language, \textit{Ul+En} additionally includes English, and \textit{All} includes all available language-specific corpora.
}
    \label{fig:language_diversity}
\end{figure}
\subsection{Effects of Increasing Language Diversity}

As shown in Figure~\ref{fig:language_diversity}, we progressively expand the search space from the user-language corpus to increasingly diverse multilingual collections to further evaluate the robustness of multilingual DCI.
The results show that as a larger number of corpora become available, the URPO gains access to complementary evidence while handling larger search spaces and more heterogeneous entity representations.
However, the DCI baseline gradually degrades as the number of accessible languages increases.
Because the baseline struggles to effectively utilize the additional corpora and becomes more susceptible to inefficient search actions.
In contrast, our method maintains stable performance and robust to increasing language diversity, which further benefits from multilingual evidence as the collection expands.

\subsection{Efficiency Analysis}
As shown in Table~\ref{tab:efficiency}, we further analyze the computational trade-off between retriever-based and DCI-based methods.
The results show that both of dense-based approaches (LcRL and D-RAG) require an additional offline stage to construct and maintain retrieval indexes, whereas DCI-based methods (\ours~and Grepseek) operate directly on raw corpora without embedding-based index construction.
The retriever-removal approach mainly shifts retrieval costs to online interactions, where the policy may need to execute multiple search commands before acquiring sufficient evidence.

In this scenario, \ours~reduces the interaction overhead while improving the accuracy of answers.
Compared with the Grepseek, the average number of search commands decreases from 4.22 to 2.56, corresponding to a 39\% reduction.
The average number of interaction turns is reduced by 27\%, from 1.87 to 1.36, while the duplicate-search rate drops from 16.4\% to 4.9\%.
More importantly, the average response length decreases from 4,853 to 3,447 tokens, a 29\% reduction, indicating a more compact interaction trajectory, which saves a lot of inference costs.
The proposed optimization does not achieve efficiency by simply truncating the search process.
Instead, it reduces redundant corpus interaction while preserving useful evidence acquisition, leading to more concise trajectories and higher task performance.

\subsection{Monolingual Compatibility}

To explore the robustness of the original capability of DCI, we further evaluate our method in the monolingual setting of GrepSeek.
We follow the original experimental setup on a multi-hop benchmark (HotpotQA), including the same Wikipedia corpus, training data, evaluation splits, and optimization configuration, and using the same token-level F1 and EM metric reported by GrepSeek.

As shown in Table~\ref{tab:mono}, although some language-specific components of our method become inapplicable when only a single corpus language is available, our method still achieves competitive performance in the original monolingual setting. 
It indicates that the proposed components preserve the original DCI capability rather than trading monolingual performance for multilingual improvements.
In addition, our method achieves the smallest discrepancy between EM and F1, further indicating that the proposed reward-driven training strategy offers more stable evaluation benefits.

\begin{table}[!t]
\centering
\small
\resizebox{\columnwidth}{!}{
\begin{tabular}{lccc}
\toprule
\textbf{HotpotQA}
& \textbf{Search-R1}
& \textbf{GrepSeek}
& \textbf{\ours}\\
\midrule
F1
& 54.24
& 62.31
& \textbf{63.51}\\
EM
& 32.02
& 46.27
& \textbf{50.00}\\
\midrule
$\Delta$ (F1-EM)
& 22.22
& 16.04
& \textbf{13.51}\\
\bottomrule
\end{tabular}
}
\caption{
Token-level F1 and EM in the original monolingual setting of GrepSeek.
}
\label{tab:mono}
\end{table}

\section{Related Work}
\paragraph{Multilingual RAG.}
Early studies on multilingual retrieval-augmented generation mainly focused on fixed retrieval-generation pipelines~\citep{chirkova-etal-2024-retrieval}, and subsequent work further improved multilingual evidence utilization by addressing language preference and cross-lingual retrieval bias~\citep{park-lee-2025-investigating,huo-etal-2026-breaking,wang-etal-2026-languages}.
More recently, agentic methods extend multilingual RAG from a fixed pipeline to iterative evidence acquisition, allowing LLMs to dynamically reformulate queries and adjust the retrieval process through multi-turn interaction~\citep{qi-etal-2026-language,lee-etal-2026-coral,10.1145/3805712.3809921}.
However, these methods still rely on conventional retrieval interfaces to access external knowledge, limiting the model's ability to directly inspect and manipulate the underlying corpus.
In contrast, our work removes this intermediate process and enables direct interaction with raw multilingual corpora, while addressing the lexical grounding and search-trajectory challenges introduced by executable search.

\paragraph{Direct Corpus Interaction.}
Recent studies have begun to rethink the retrieval interface itself through direct corpus interaction (DCI), where language agents operate directly on raw corpora through executable search commands instead of accessing evidence through a fixed similarity-based retriever~\citep{li2026semanticsimilarityrethinkingretrieval,salemi2026grepseektrainingsearchagents}.
This paradigm provides finer-grained control over evidence acquisition and enables agents to learn effective corpus interaction through multi-step search and reinforcement learning.
However, existing DCI studies primarily focus on monolingual settings, where search commands operate over corpora with relatively consistent lexical forms.
Multilingual DCI introduces additional challenges because semantically equivalent entities may be expressed differently across languages, while the search policy must coordinate its trajectory over language-specific corpora.
Our work extends direct corpus interaction to multilingual evidence acquisition and explicitly addresses these challenges through corpus-native grounding and redundancy-aware policy optimization.

\section{Conclusion}

In this work, we propose \ours, a retriever-free multilingual search framework that enables LLMs to acquire evidence directly from language-specific corpora through executable commands.
We introduce Utility-Aware Retrieval Policy Optimization with corpus-native entity grounding and redundancy-aware search reward, which improves lexical evidence access and promotes more effective search trajectories.
Experimental results demonstrate that \ours~consistently outperforms strong retrieval-based and DCI baselines, generalizes effectively to unseen languages, and remains robust as the multilingual search space expands.
Moreover, \ours~substantially reduces redundant corpus interactions while preserving strong monolingual DCI capability.

\section*{Limitations}

Like most post-training approaches, \ours~requires task-specific training data and additional optimization cost.
Due to these limitations, we conducted tests on three tasks in different languages, which only covers the surface of the world’s languages. 
In the future, it will be appropriate to study the generality of our framework compared to other LLMs.
Although corpus-native grounding alleviates cross-lingual surface-form mismatch, direct corpus interaction remains inherently sensitive to lexical variation, especially when useful aliases are not explicitly exposed by the corpus.
Future work may explore hybrid retrieval that combines executable lexical search with semantic retrieval, preserving precise corpus control while improving robustness to surface-form variation.

\section*{Ethics Statement}
All data used in our experiments were obtained exclusively from open and publicly available benchmarks. Statistics related to language distributions in commonly used pre-training datasets were sourced from official open-resourced technical report. Our analysis and methodology were conducted with sensitivity, and do not involve or affect attributes related to gender, sex, or race. Therefore, in this study, we did not explicitly encounter any ethical statements.


\bibliography{custom}

\clearpage
\appendix

\section{Model Details}
\label{app:baselines}
We compare our method with various representative baselines in multilingual RAG. The baselines are as follows:

\paragraph{Direct Inference} directly prompts the backbone LLMs to answer each question without access to external knowledge.

\paragraph{Naive RAG} follows the conventional single-turn RAG pipeline, where a multilingual retriever retrieves top-ranked passages from the multilingual Wikipedia collections before answer generation.

\paragraph{D-RAG}~\citep{ranaldi-etal-2025-improving-multilingual} enhances multilingual RAG with dialectic reasoning over retrieved evidence, explicitly identifying and reconciling relevant information before answer generation.

\paragraph{LcRL}~\citep{qi-etal-2026-language} is a multilingual search-augmented RL method that jointly optimizes multilingual retrieval behavior and generation through language-coupled rollout and reward modeling.

\paragraph{Naive DCI}~\citep{li2026semanticsimilarityrethinkingretrieval} directly exposes the multilingual corpora to the backbone model through executable search commands, without task-specific post-training, measuring the intrinsic ability of the model to perform multilingual direct corpus interaction.

\paragraph{GrepSeek}~\citep{salemi2026grepseektrainingsearchagents} initializes the model with direct corpus interaction training and optimizes executable search behavior with its original learning objective. We adapt its corpus interface to the same multilingual collections while keeping its training formulation unchanged.

\paragraph{\ours} uses the same direct corpus interaction environment and initialization as GrepSeek, while introducing corpus-native entity grounding and utility-aware retrieval policy optimization with RL.

\section{Specific Results for Each Language}
\label{app:detail_results}
We report the specific performance of \ours~in all languages on MKQA and MLQA. The results show that \ours~with URPO consistently outperform all baselines in nearly all languages. 
Several language-specific patterns can be observed.

Among retrieval-based multilingual methods, the agentic baseline LcRL shows relatively limited EM performance, especially on MLQA.
This may partly result from the sparsity of exact-match rewards, which provide weak optimization signals for multi-turn retrieval policies.
In contrast, \ours~is also trained with EM-based GRPO but achieves substantially stronger EM, indicating that direct corpus interaction combined with our utility-aware optimization is more effective under sparse exact-match supervision.

Within the DCI paradigm, the training-free baseline exhibits clear language sensitivity.
It performs relatively well on languages such as English and French, but degrades markedly on Chinese, Japanese, and Arabic.
\ours~ and GrepSeek consistently improves over this baseline, showing that post-training is important for maintaining effective corpus-interaction behavior across languages.

Moreover, compared with the training-free DCI baseline, the improvements are particularly pronounced on languages where DCI performs poorly, such as Chinese, Japanese, Arabic, and Korean.
These languages tend to involve larger cross-lingual surface-form variation, making exact lexical search more vulnerable to entity mismatch.
The larger gains therefore further support the role of corpus-native grounding in reducing language-specific lexical barriers during multilingual evidence acquisition.

\begin{table*}[ht]
    \centering
    \small

    \definecolor{colorMethods}{RGB}{255,255,255}
    \definecolor{colorSeen}{RGB}{226,243,246}
    \definecolor{colorUnseen}{RGB}{238,228,241}
    \definecolor{colorAve}{RGB}{250,235,215}
\resizebox{\textwidth}{!}{
    \begin{tabular}{
        >{\columncolor{colorMethods}}l
        *{8}{>{\columncolor{colorSeen}}c}
        >{\columncolor{colorAve}}c
        *{5}{>{\columncolor{colorUnseen}}c}
        >{\columncolor{colorAve}}c
    }
        \toprule
        \rowcolor{colorMethods}
        \multirow{2}{*}{\textbf{Methods}}
        & \multicolumn{8}{c}{\textbf{MKQA (Seen Languages)}}
        & \textbf{}
        & \multicolumn{5}{c}{\textbf{MKQA (Unseen Languages)}}
        & \textbf{} \\
        \cmidrule(lr){2-9}
        \cmidrule(lr){11-15}
        \rowcolor{colorMethods}
        & \textbf{En}
        & \textbf{Fr}
        & \textbf{It}
        & \textbf{Ru}
        & \textbf{Fi}
        & \textbf{Zh}
        & \textbf{Ja}
        & \textbf{Ar}
        & \textbf{Avg.}
        & \textbf{Es}
        & \textbf{Pt}
        & \textbf{Ko}
        & \textbf{Th}
        & \textbf{De}
        & \textbf{Avg.}\\
        \midrule

        \multicolumn{16}{c}{
            \cellcolor{colorMethods}
            \textbf{\texttt{Qwen3.5-4B}}
        } \\
        \midrule

Direct
& 24.3 & 18.6 & 17.2 & 12.7 & 12.3 & 11.9 & 6.8 & 7.1
& 13.9
& 18.6 & 17.4 & 7.1 & 6.3 & 19.2
& 13.7 \\

RAG
& 41.7 & 31.6 & 30.5 & 23.8 & 22.2 & 11.7 & 17.8 & 12.9
& 24.0
& 30.1 & 29.6 & 10.4 & 11.8 & 27.8
& 21.9 \\

D-RAG
& 43.9 & 33.7 & 28.5 & \underline{24.2} & 29.4 & 19.3 & 20.6 & \underline{17.5}
& 27.1
& 35.1 & 14.8 & 13.5 & 10.7 & \underline{42.2}
& 23.3 \\

LcRL
& 36.1 & 30.6 & 28.9 & 21.9 & 27.1 & 16.2 & 19.0 & 12.0
& 24.0
& 27.7 & 29.1 & 12.1 & 9.2 & 26.8
& 21.0 \\

DCI
& 45.0 & 38.4 & 35.2 & 19.0 & 34.1 & 5.0 & 8.3 & 7.0
& 24.0
& 38.1 & 36.6 & 6.6 & 12.0 & 38.2
& 26.3 \\

Grepseek
& \underline{48.2} & \underline{41.9} & \underline{41.9} & \underline{30.9} & \underline{35.0} & \underline{24.8} & \underline{30.7} & 17.8
& \underline{33.9}
& \underline{40.6} & \underline{38.5} & \underline{20.4} & \textbf{13.1} & 39.9
& \underline{30.5} \\

\ours
& \textbf{52.0} & \textbf{46.2} & \textbf{44.1} & \textbf{33.0} & \textbf{37.1} & \textbf{26.5} & \textbf{32.3} & \textbf{19.3}
& \textbf{36.3}
& \textbf{44.9} & \textbf{43.1} & \textbf{21.8} & \underline{13.0} & \textbf{43.2}
& \textbf{33.2} \\

        \midrule

        \multicolumn{16}{c}{
            \cellcolor{colorMethods}
            \textbf{\texttt{Qwen3.5-9B}}
        } \\
        \midrule

Direct
& 32.1 & 23.6 & 22.6 & 16.4 & 16.6 & 13.5 & 10.5 & 9.8
& 18.1
& 25.5 & 23.6 & 9.0 & 9.3 & 23.5
& 18.2 \\

RAG
& 44.9 & 33.8 & 31.5 & 24.7 & 24.1 & 12.6 & 19.2 & 12.2
& 25.4
& 33.0 & 31.1 & 12.4 & 13.9 & 29.2
& 23.9 \\

D-RAG
& 49.4 & 34.9 & 32.5 & 24.2 & 26.8 & 14.7 & 21.6 & 12.5
& 27.1
& 34.2 & 24.6 & 12.6 & 9.5 & 28.5
& 21.9 \\

LcRL
& 38.7 & 33.7 & 32.0 & 25.4 & 30.4 & 20.1 & 22.4 & 15.4
& 27.2
& 36.3 & 31.0 & 15.4 & 11.8 & 33.7
& 25.6 \\

DCI
& 51.6 & 42.6 & 41.7 & 17.7 & \textbf{43.5} & 3.2 & 6.5 & 7.4
& 26.8
& 40.9 & 40.4 & 3.4 & 15.9 & \underline{44.2}
& 29.0 \\

Grepseek
& \underline{51.7} & \underline{44.6} & \underline{44.7} & \underline{36.6} & 36.6 & \underline{24.8} & \underline{35.1} & \underline{22.5}
& \underline{37.1}
& \underline{45.3} & \underline{43.2} & \underline{26.9} & \underline{18.9} & 43.6
& \underline{35.6} \\

\ours
& \textbf{53.6} & \textbf{49.4} & \textbf{46.8} & \textbf{38.7} & \underline{42.5} & \textbf{31.2} & \textbf{38.1} & \textbf{24.8}
& \textbf{40.6}
& \textbf{48.6} & \textbf{45.2} & \textbf{29.1} & \textbf{19.3} & \textbf{45.6}
& \textbf{37.6} \\

        \bottomrule
    \end{tabular}
}
    \caption{
    Main results of EM (\%) on MKQA. Background colors distinguish \colorbox{colorSeen}{seen languages}, \colorbox{colorUnseen}{unseen languages}, and \colorbox{colorAve}{average performance}.
    The highest score under each base model is highlighted in \textbf{bold}, and the second-highest score is \underline{underlined}.
    }
    \label{tab:mkqa_results_em}
\end{table*}

\begin{table*}[ht]
    \centering
    \small

    \definecolor{colorMethods}{RGB}{255,255,255}
    \definecolor{colorSeen}{RGB}{226,243,246}
    \definecolor{colorUnseen}{RGB}{238,228,241}
    \definecolor{colorAve}{RGB}{250,235,215}
\resizebox{\textwidth}{!}{
    \begin{tabular}{
        >{\columncolor{colorMethods}}l
        *{8}{>{\columncolor{colorSeen}}c}
        >{\columncolor{colorAve}}c
        *{5}{>{\columncolor{colorUnseen}}c}
        >{\columncolor{colorAve}}c
    }
        \toprule
        \rowcolor{colorMethods}
        \multirow{2}{*}{\textbf{Methods}}
        & \multicolumn{8}{c}{\textbf{MKQA (Seen Languages)}}
        & \textbf{}
        & \multicolumn{5}{c}{\textbf{MKQA (Unseen Languages)}}
        & \textbf{} \\
        \cmidrule(lr){2-9}
        \cmidrule(lr){11-15}
        \rowcolor{colorMethods}
        & \textbf{En}
        & \textbf{Fr}
        & \textbf{It}
        & \textbf{Ru}
        & \textbf{Fi}
        & \textbf{Zh}
        & \textbf{Ja}
        & \textbf{Ar}
        & \textbf{Avg.}
        & \textbf{Es}
        & \textbf{Pt}
        & \textbf{Ko}
        & \textbf{Th}
        & \textbf{De}
        & \textbf{Avg.}\\
        \midrule

        \multicolumn{16}{c}{
            \cellcolor{colorMethods}
            \textbf{\texttt{Qwen3.5-4B}}
        } \\
        \midrule

Direct
& 29.6 & 22.7 & 20.4 & 15.0 & 14.3 & 17.4 & 10.5 & 8.4
& 17.3
& 23.0 & 21.5 & 9.2 & 7.8 & 23.4
& 17.0 \\

RAG
& 53.6 & 39.6 & 36.8 & 29.4 & 28.8 & 16.6 & 24.6 & 16.8
& 30.8
& 37.7 & 37.7 & 13.7 & 14.3 & 34.8
& 27.6 \\

D-RAG
& 49.7 & 39.5 & 34.8 & 30.3 & 35.7 & 25.5 & 26.9 & 23.7
& 33.3
& 41.2 & 20.2 & 19.7 & 15.9 & 48.4
& 29.1 \\

LcRL
& 44.2 & 36.4 & 33.9 & 25.1 & 31.8 & 22.0 & 22.9 & 14.9
& 28.9
& 39.4 & 37.4 & 14.0 & 13.0 & 31.6
& 27.1 \\

DCI
& 60.3 & 53.7 & 51.1 & 32.9 & 47.5 & 18.8 & 25.4 & 14.4
& 38.0
& 53.8 & 50.3 & 18.8 & 19.8 & 53.1
& 39.2 \\

Grepseek
& \underline{68.7} & \underline{59.8} & \underline{58.2} & \underline{41.3} & \underline{49.6} & \underline{34.5} & \underline{41.8} & \underline{26.2}
& \underline{47.5}
& \underline{58.1} & \underline{54.4} & \underline{30.0} & \underline{27.1} & \underline{57.7}
& \underline{45.5} \\

\ours
& \textbf{73.4} & \textbf{63.7} & \textbf{61.2} & \textbf{46.1} & \textbf{53.6} & \textbf{36.0} & \textbf{44.6} & \textbf{30.6}
& \textbf{51.2}
& \textbf{62.3} & \textbf{58.8} & \textbf{33.5} & \textbf{28.6} & \textbf{62.6}
& \textbf{49.1} \\

        \midrule

        \multicolumn{16}{c}{
            \cellcolor{colorMethods}
            \textbf{\texttt{Qwen3.5-9B}}
        } \\
        \midrule

Direct
& 39.9 & 30.4 & 27.3 & 19.7 & 20.1 & 22.4 & 15.9 & 13.0
& 23.6
& 31.2 & 28.7 & 14.3 & 12.8 & 29.4
& 23.3 \\

RAG
& 56.8 & 43.2 & 39.2 & 32.1 & 31.9 & 19.5 & 26.5 & 17.3
& 33.3
& 41.7 & 40.5 & 18.0 & 19.7 & 37.7
& 31.5 \\

D-RAG
& 55.5 & 40.7 & 38.2 & 30.8 & 32.6 & 20.4 & 27.6 & 18.2
& 33.0
& 40.5 & 30.4 & 18.9 & 15.4 & 34.7
& 28.0 \\

LcRL
& 47.1 & 39.9 & 37.0 & 29.1 & 35.1 & 26.0 & 26.6 & 18.7
& 32.4
& 46.1 & 37.8 & 17.2 & 17.5 & 35.1
& 30.7 \\

DCI
& \underline{73.6} & \underline{63.8} & \underline{64.7} & 37.7 & \textbf{60.1} & 20.5 & 30.9 & 19.0
& 46.3
& 63.3 & \underline{61.6} & 20.8 & 28.7 & \textbf{65.6}
& 48.0 \\

Grepseek
& 72.3 & 62.7 & 64.1 & \underline{50.2} & 54.6 & \underline{38.9} & \underline{48.3} & \underline{34.7}
& \underline{53.2}
& \underline{64.2} & 60.8 & \underline{39.4} & \underline{32.9} & \underline{65.2}
& \underline{52.5} \\

\ours
& \textbf{74.2} & \textbf{66.3} & \textbf{65.7} & \textbf{50.7} & \underline{58.2} & \textbf{43.9} & \textbf{50.0} & \textbf{35.2}
& \textbf{55.5}
& \textbf{67.1} & \textbf{62.0} & \textbf{41.0} & \textbf{35.2} & 65.1
& \textbf{54.1} \\

        \bottomrule
    \end{tabular}
}
    \caption{
    Main results of fEM (\%) on MKQA. Background colors distinguish \colorbox{colorSeen}{seen languages}, \colorbox{colorUnseen}{unseen languages}, and \colorbox{colorAve}{average performance}.
    The highest score under each base model is highlighted in \textbf{bold}, and the second-highest score is \underline{underlined}.
    }
    \label{tab:mkqa_results_fem}
\end{table*}

\begin{table*}[!t]
    \centering
    \small
\resizebox{\textwidth}{!}{
    \begin{tabular}{
        lcccccccccccccccc
    }
        \toprule
        \multirow{2}{*}{\textbf{Methods}}
        & \multicolumn{7}{c}{\textbf{EM}}
        & \textbf{}
        & \multicolumn{7}{c}{\textbf{fEM}}
        & \textbf{} \\
        \cmidrule(lr){2-8}
        \cmidrule(lr){10-16}

        & \textbf{En}
        & \textbf{De}
        & \textbf{Es}
        & \textbf{Hi}
        & \textbf{Vi}
        & \textbf{Zh}
        & \textbf{Ar}
        & \textbf{Avg.}
        & \textbf{En}
        & \textbf{De}
        & \textbf{Es}
        & \textbf{Hi}
        & \textbf{Vi}
        & \textbf{Zh}
        & \textbf{Ar}
        & \textbf{Avg.} \\
        \midrule

        \multicolumn{17}{c}{
            \textbf{\texttt{Qwen3.5-4B}}
        } \\
        \midrule

        Direct
& 9.4 & 6.6 & 5.5 & 2.1 & 6.6 & 4.1 & 2.5
& 5.3
& 11.0 & 7.7 & 6.1 & 2.3 & 7.8 & 5.0 & 3.1
& 6.1 \\

RAG
& 14.7 & 11.6 & 9.3 & 10.1 & 14.4 & 6.3 & 6.6
& 10.4
& 18.3 & 14.0 & 11.2 & 11.8 & 16.1 & 8.7 & 8.6
& 12.7 \\

D-RAG
& \textbf{17.2} & 13.2 & 9.8 & 10.7 & \underline{15.1} & 7.6 & 7.4
& 11.6
& 19.8 & 15.8 & 12.6 & 13.5 & 17.8 & 10.2 & 10.2
& 14.3 \\

LcRL
& 12.4 & 8.7 & 8.2 & 5.5 & 7.6 & 7.2 & 3.3
& 7.6
& 14.7 & 11.4 & 13.9 & 6.4 & 10.0 & 9.6 & 4.5
& 10.1 \\

DCI
& 14.9 & 12.1 & 11.8 & 3.9 & 12.9 & 8.9 & 5.4
& 10.0
& 22.9 & 18.9 & 21.5 & 9.2 & 20.7 & 18.6 & 9.9
& 17.4 \\

Grepseek
& 14.5 & \underline{13.7} & \underline{11.9} & \underline{15.8} & 15.0 & \underline{13.3} & \underline{13.0}
& \underline{13.9}
& \textbf{34.0} & \underline{32.1} & \textbf{30.9} & \underline{29.4} & \underline{44.1} & \underline{28.1} & \underline{23.7}
& \underline{31.7} \\

\ours
& \underline{16.2} & \textbf{14.5} & \textbf{14.7} & \textbf{16.8} & \textbf{21.9} & \textbf{14.4} & \textbf{13.7}
& \textbf{16.0}
& \underline{33.2} & \textbf{33.9} & \underline{30.6} & \textbf{33.1} & \textbf{45.7} & \textbf{30.9} & \textbf{27.5}
& \textbf{33.6} \\

        \midrule

        \multicolumn{17}{c}{
            \textbf{\texttt{Qwen3.5-9B}}
        } \\
        \midrule

        Direct
& 9.8 & 7.7 & 7.1 & 2.9 & 7.3 & 5.2 & 3.0
& 6.1
& 12.0 & 9.5 & 8.5 & 3.9 & 8.2 & 6.6 & 3.8
& 7.5 \\

RAG
& 16.2 & 11.7 & 10.5 & 12.0 & 13.6 & 7.8 & 6.1
& 11.1
& 19.7 & 14.8 & 13.0 & 13.7 & 17.6 & 11.1 & 8.7
& 14.1 \\

D-RAG
& 17.5 & 13.6 & 11.9 & 12.9 & 15.4 & 9.8 & 7.9
& 12.7
& 20.2 & 16.3 & 14.6 & 15.3 & 18.1 & 12.6 & 10.4
& 15.3 \\

LcRL
& 13.6 & 11.4 & 10.4 & 7.0 & 10.8 & 11.6 & 6.2
& 10.1
& 16.8 & 13.1 & 17.0 & 11.8 & 13.3 & 13.3 & 7.2
& 13.2 \\

DCI
& 17.0 & 13.4 & 11.0 & 6.3 & 17.1 & 7.6 & 5.9
& 11.2
& 25.0 & 20.9 & 20.3 & 14.9 & 30.6 & 21.7 & 16.1
& 21.4 \\

Grepseek
& \underline{18.5} & \underline{15.0} & \underline{15.0} & \underline{15.5} & \underline{20.6} & \underline{12.5} & \underline{8.1}
& \underline{15.0}
& \underline{37.7} & \textbf{34.3} & \underline{30.3} & \underline{30.1} & \underline{44.7} & \underline{29.4} & \underline{25.6}
& \underline{33.1} \\

\ours
& \textbf{19.2} & \textbf{15.2} & \textbf{16.2} & \textbf{16.7} & \textbf{22.2} & \textbf{13.5} & \textbf{12.4}
& \textbf{16.5}
& \textbf{38.0} & \underline{34.1} & \textbf{35.7} & \textbf{32.0} & \textbf{46.6} & \textbf{30.6} & \textbf{27.3}
& \textbf{34.9} \\

        \bottomrule
    \end{tabular}}

    \caption{
    Main results (\%) on MLQA measured by EM and fEM. The highest score under each base model is highlighted in \textbf{bold}, and the second-highest score is \underline{underlined}.
    }
    \label{tab:main_mlqa}
\end{table*}

\begin{table*}[ht]
    \centering
    \small

    \begin{tabular}{llccccccccc
    }
        \toprule

        \textbf{Methods}
        & \textbf{Corpora} 
        & \textbf{En}
        & \textbf{Fr}
        & \textbf{It}
        & \textbf{Ru}
        & \textbf{Fi}
        & \textbf{Zh}
        & \textbf{Ja}
        & \textbf{Ar}
        & \textbf{Avg.}\\
        \midrule


        \multirow{3}{*}{Grepseek} & Ul
        & 40.6 & 37.5 & 37.1 & 31.3 & 33.1 & 27.1 & 29.4 & 22.4 & 32.3 \\

        & Ul+En
        & 53.6 & 48.9 & 45.2 & 37.4 & 39.8 & 28.8 & 36.9 & 22.9 & 39.2 \\

        & All
       & 51.7 & 44.6 & 44.7 & 36.6 & 36.6 & 24.8 & 35.1 & 22.5 & 37.1 \\
        \midrule
        \multirow{3}{*}{\ours} & Ul
        & 51.9 & 46.3 & 42.4 & 34.8 & 36.7 & 25.5 & 37.4 & 21.4 & 37.1 \\

        & Ul+En
        & 53.0 & 46.2 & 46.0 & 37.9 & 42.0 & 31.6 & 37.7 & 23.9 & 39.8 \\

        & All
        & 53.6 & 49.4 & 46.8 & 38.7 & 42.5 & 31.2 & 38.1 & 24.8 & 40.6 \\

        \bottomrule
    \end{tabular}
    \caption{
    Results of EM (\%) on MKQA for corpus-language coverage, based on \texttt{Qwen3.5-9B}. \textit{Ul} denotes the corpus in the user-query language, \textit{Ul+En} additionally includes English, and \textit{All} includes all available language-specific corpora.
    The highest score under each base model is highlighted in \textbf{bold}, and the second-highest score is \underline{underlined}.
    }
    \label{tab:mkqa_language_results}
\end{table*}

\section{Training Template}
\label{sec:template}
To enable the policy model to autonomously navigate multilingual corpora, we employ a structured template that establishes a unified interaction protocol between the LLM and the corpus environment.
As shown in Table~\ref{tab:system_prompt}, each trajectory alternates among reasoning in \texttt{<think>}, executable corpus search in \texttt{<tool\_call>}, and final answer generation in \texttt{<answer>}.
The template exposes available language-specific corpora and shell operations, while allowing the policy to determine the target language and search command at each turn.
Corpus observations are returned through \texttt{<tool\_response>} and may contain \texttt{[anchor]} or \texttt{[active-anchor]} fields that provide corpus-observed entity forms for subsequent search.

\definecolor{multicolor}{RGB}{70, 120, 180}
\newcommand{\multihighlight}[1]{\textcolor{multicolor}{\textbf{#1}}}

\begin{table*}[!t]
\centering

\begin{tcolorbox}[
    colframe=gray!50,
    arc=4pt,
    boxrule=0.5pt,
    left=10pt,
    right=10pt,
    top=10pt,
    bottom=10pt
]
\small

\textbf{System Prompt:} \\

You are a research agent that answers questions by searching a
\multihighlight{MULTILINGUAL Wikipedia corpus collection}
using shell commands.
The corpus is split by language and accessed through
\texttt{corpus/<lang>}, e.g.,
\texttt{corpus/en}, \texttt{corpus/zh}, and \texttt{corpus/ja}.
Always cap retrieved output with \texttt{head}. \\

\textbf{Cross-lingual Search Strategy:} \\

You decide \multihighlight{which language-specific corpus to search}.
The question language is a useful starting point, but relevant evidence may
be better covered in \multihighlight{another language's Wikipedia}.
You may search multiple language-specific corpora when needed. \\

\multihighlight{Translate or transliterate key entities across languages}
when appropriate, and use the entity form most likely to appear in the
target language corpus. \\

The tool response may contain an
\texttt{[anchor]} field with
\multihighlight{corpus-observed entity titles or aliases}.
These forms may be reused as candidate search keys in subsequent interactions.
When available, \texttt{[active-anchor]} denotes an
\multihighlight{entity form recovered after an empty lexical search}. \\

Allowed shell tools include
\texttt{rg}, \texttt{grep}, \texttt{find}, \texttt{sed},
\texttt{awk}, \texttt{head}, \texttt{tail}, \texttt{cat},
\texttt{wc}, \texttt{sort}, \texttt{cut}, \texttt{uniq}, and \texttt{tr}.
Pipes are allowed, while redirection, command chaining,
and command substitution are not allowed. \\

\textbf{Interaction Format:} \\

For every step, first write a short paragraph of reasoning in plain prose
(2--5 sentences), describing what has been learned, what is still missing,
and \multihighlight{which language corpus should be searched next and why}.
Then output exactly one of the following: \\

{\color{tagpurple}\texttt{<tool\_call>}} \\
\texttt{\{"name": "shell", "arguments": \{"command": "one or more command lines"\}\}} \\
{\color{tagpurple}\texttt{</tool\_call>}} \\

\textbf{or}, only when sufficient evidence has been obtained: \\

{\color{tagorange}\texttt{<answer>}}
\texttt{the final answer}
{\color{tagorange}\texttt{</answer>}} \\

Always reason first and then produce exactly one action block.
The final answer should be concise and written in
\multihighlight{the same language as the question}.

\vspace{5pt}
\hrule
\vspace{5pt}

\textbf{Question:} \{question\}

\end{tcolorbox}

\caption{System prompt for \ours. Multilingual-specific instructions are highlighted in blue.}
\label{tab:system_prompt}

\end{table*}

\section{Case Study}
\label{sec:case_study}
\begin{CJK*}{UTF8}{gkai}
We present two representative cases to illustrate how corpus-native grounding facilitates multilingual evidence acquisition.
\textit{First}, figure~\ref{fig:case_anchor} shows the standard grounding process: the model first retrieves Chinese evidence that exposes the English alias \textit{Beauty and the Beast}, then reuses this corpus-observed form to search the English collection and identify the correct answer. 
\textit{Second}, figure~\ref{fig:case_active_anchor} demonstrates grounding under empty retrieval.
When the original title \textit{I Just Can't Get Enough} fails to match the Chinese corpus, Active-Anchor uses the English pivot corpus to recover the corpus-native title \textit{Just Can't Get Enough}.
The recovered form is then reused to successfully retrieve the relevant Chinese evidence and answer the question. 
Overall, these cases show that our grounding mechanism improves multilingual evidence access by converting corpus-observed entity forms into reusable search keys, including both cross-lingual alias reuse and recovery from failed lexical matches.

\definecolor{hlorangecolor}{RGB}{252, 232, 205}
\definecolor{anchorred}{RGB}{248, 220, 220}
\newcommand{\hlanchor}[1]{%
    \bgroup
    \markoverwith{\textcolor{anchorred}{\rule[-0.6ex]{2pt}{2.6ex}}}%
    \ULon{#1}%
}
\newcommand{\hlorange}[1]{%
    \bgroup
    \markoverwith{\textcolor{hlorangecolor}{\rule[-0.6ex]{2pt}{2.6ex}}}%
    \ULon{#1}%
}

\begin{figure*}[ht]
    \centering
    \small
    \begin{tcolorbox}[
        colback=white,
        colframe=boxborder,
        boxrule=1.5pt,
        arc=0pt,
        left=4pt, right=4pt, top=4pt, bottom=4pt,
        fonttitle=\bfseries
    ]

        \begin{tcolorbox}[
            colback=headergray,
            colframe=headergray,
            sharp corners,
            boxsep=0pt, left=2pt, right=2pt, top=2pt, bottom=2pt
        ]
            \textbf{Question:} 美女与野兽动画片里 Belle 的扮演者是谁？ \\
            \textbf{Golden Answer:} 佩吉·奥哈拉（Paige O'Hara）
        \end{tcolorbox}

        \textbf{Round 1} \\
        {\color{tagblue}\texttt{<think>}}
        \hlblue{问题询问动画电影《美女与野兽》中 Belle 的演员。
        我先在中文语料库中定位《美女与野兽》和 Belle 对应的角色信息。}
        {\color{tagblue}\texttt{</think>}} \\
        \begin{center}
            \vspace{-5pt}
            \textcolor{orangetext}{\textit{\textbf{Translate "Belle" and Search in the Chinese collection.}}} \\
            \textcolor{arrowcolor}{\scalebox{1.5}{$\downarrow$}}
            \vspace{-5pt}
        \end{center}

        {\color{tagpurple}\texttt{<tool\_call>}}
        \hlpurple{\texttt{rg -F "美女与野兽" corpus/zh | rg -F "贝儿" | head -n 5}}
        {\color{tagpurple}\texttt{</tool\_call>}} \\
        
        {\color{taggreen}\texttt{<tool\_response>}} \\
        \textit{[lang=zh]} \\
        \texttt{[ZH]} Doc 1 :
        …\hlgreen{贝儿}是\hlgreen{《美女与野兽》}中的主要角色之一，并与其他迪士尼公主一同出现在多部衍生作品中。 \\[2pt]
        \texttt{[ZH]} Doc 2:
         该作为1991年迪士尼动画电影
        \hlgreen{《美女与野兽》} \hlanchor{（Beauty and the Beast）}
        的衍生故事，故事发生在\hlgreen{贝儿}被困于城堡期间。 \\[2pt]
        \texttt{[ZH]} Doc 3 : 
         昆夏美曾为电影\hlgreen{《美女与野兽》}的主角\hlgreen{贝儿}担任日语版配音。 \\
        \vspace{2pt}
        \hlanchor{\textbf{\texttt{[anchor]} 
        title=\texttt{"美女与野兽"} 
        $\mid$ aka: \texttt{Beauty and the Beast}}} \\

        {\color{taggreen}\texttt{</tool\_response>}} \\

        \textbf{Round 2} \\
        {\color{tagblue}\texttt{<think>}}
        \hlblue{检索结果表明《美女与野兽》的英文名称是
        ``Beauty and the Beast''。
        我可以复用该语料库中观察到的英文 \\
        实体形式，并结合问题中已有的
        ``Belle''，在英文语料库中进一步检索原版动画的配音信息。}
        {\color{tagblue}\texttt{</think>}} \\

        \begin{center}
            \vspace{-5pt}
            \textcolor{orangetext}{\textit{\textbf{Reuse the grounded alias in the English collection.}}} \\
            \textcolor{arrowcolor}{\scalebox{1.5}{$\downarrow$}}
            \vspace{-5pt}
        \end{center}


        {\color{tagpurple}\texttt{<tool\_call>}}
        \hlpurple{\texttt{rg -F "Beauty and the Beast" corpus/en | rg -F "Belle" | head -n 5}}
        {\color{tagpurple}\texttt{</tool\_call>}} \\

        {\color{taggreen}\texttt{<tool\_response>}} \\
        \textit{[lang=en]} \\
        \texttt{[EN]} Doc 1 : \hlorange{Paige O'Hara} as \hlgreen{Belle}, a bright young woman who seeks adventure
        and offers her own freedom to the Beast in return for her father's... \\[2pt]
        \texttt{[EN]} Doc 2 : \hlgreen{Belle} is a fictional character in Disney's \hlgreen{Beauty and the Beast} franchise.
        She first appeared in the 1991 animated film and was
        \hlorange{voiced by Paige O'Hara}. \\[2pt]
        \texttt{[EN]} Doc 3 : Disney's 1991 animated classic features \hlgreen{Belle} and the Beast.
        The cast includes \hlorange{Paige O'Hara}, Robby Benson,
        Angela Lansbury, and Jerry Orbach... \\

        {\color{taggreen}\texttt{</tool\_response>}} \\


        {\color{tagblue}\texttt{<think>}}
        \hlblue{多个英文文档均表明，1991年迪士尼动画电影
        《美女与野兽》中 Belle 的配音演员是 Paige O'Hara。}
        {\color{tagblue}\texttt{</think>}} \\

        {\color{tagorange}\texttt{<answer>}}
        \hlorange{佩吉·奥哈拉（Paige O'Hara）}。
        {\color{tagorange}\texttt{</answer>}}

    \end{tcolorbox}

    \caption{Case study of corpus-native entity grounding.
    The model first retrieves Chinese evidence that exposes the
    corpus-observed English title \textit{Beauty and the Beast},
    and then reuses this grounded alias to acquire the answer
    from the English collection. Green blocks denote useful
    evidence or grounded entity forms.}
    \label{fig:case_anchor}
\end{figure*}

\begin{figure*}[ht]
    \centering
    \small
    \begin{tcolorbox}[
        colback=white,
        colframe=boxborder,
        boxrule=1.5pt,
        arc=0pt,
        left=4pt, right=4pt, top=4pt, bottom=4pt,
        fonttitle=\bfseries
    ]

        \begin{tcolorbox}[
            colback=headergray,
            colframe=headergray,
            sharp corners,
            boxsep=0pt, left=2pt, right=2pt, top=2pt, bottom=2pt
        ]
            \textbf{Question:} 《I Just Can't Get Enough》是谁演唱的？ \\
            \textbf{Golden Answer:} 流行尖端（Depeche Mode）
        \end{tcolorbox}

        \textbf{Round 1} \\

        {\color{tagblue}\texttt{<think>}}
        \hlblue{问题询问歌曲《I Just Can't Get Enough》的演唱者。
        我先使用问题中给出的歌曲名称，在中文语料库中\\进行精确检索。}
        {\color{tagblue}\texttt{</think>}} \\

        \begin{center}
            \vspace{-5pt}
            \textcolor{orangetext}{\textit{\textbf{Search the queried title in the Chinese collection.}}} \\
            \textcolor{arrowcolor}{\scalebox{1.5}{$\downarrow$}}
            \vspace{-5pt}
        \end{center}

        {\color{tagpurple}\texttt{<tool\_call>}}
        \hlpurple{\texttt{rg -F "I Just Can't Get Enough" corpus/zh | head -n 5}}
        {\color{tagpurple}\texttt{</tool\_call>}} \\

        {\color{taggreen}\texttt{<tool\_response>}} \\
        \textit{[lang=zh]} \\
        \texttt{[no results]} \\[-1pt]
        \vspace{2pt}
        
        \hlanchor{\textbf{\textit{[active-anchor: fallback to English pivot]}}} \\
        {\color{taggreen}\texttt{</tool\_response>}} \\
        
        {\color{tagpurple}\texttt{<tool\_call>}}
        \hlpurple{\texttt{rg -F "I Just Can't Get Enough" corpus/en | head -n 5}}
        {\color{tagpurple}\texttt{</tool\_call>}} \\

        {\color{taggreen}\texttt{<tool\_response>}} \\
        \texttt{[EN]} Doc 1 :
        \quad \hlanchor{``Just Can't Get Enough''} is a 1981 song by the English electronic band Depeche Mode. … “And \hlgreen{I just can't get enough}, And I just can't get enough…" \\[2pt]
        \texttt{[EN]} Doc 2 :
        \quad The song was released in 1981 as a single from Depeche Mode's debut album
        \textit{Speak \& Spell}. The refrain begins with
        \hlgreen{``I just can't get enough''}... \\[2pt]
        \texttt{[EN]} Doc 3 :
        \quad The album includes the single
        \hlanchor{``Just Can't Get Enough''}, one of the band's early hits... \\
        \vspace{2pt}
        \hlanchor{\textbf{\texttt{[active-anchor]}
        title=\texttt{"I Just Can't Get Enough"}
        $\mid$ aka: \texttt{Just Can't Get Enough}}} \\
        {\color{taggreen}\texttt{</tool\_response>}} \\

        \textbf{Round 2} \\

        {\color{tagblue}\texttt{<think>}}
        \hlblue{中文语料库中没有与 ``I Just Can't Get Enough'' 完全匹配的结果。
        Active-Anchor 在英文枢纽语料库中发现，该表达对应的规范歌曲标题是
        ``Just Can't Get Enough''。我将复用这一 corpus-native 名称重新检索中文语料库。}
        {\color{tagblue}\texttt{</think>}} \\

        \begin{center}
            \vspace{-5pt}
            \textcolor{orangetext}{\textit{\textbf{Reuse the recovered corpus-native title in the Chinese collection.}}} \\
            \textcolor{arrowcolor}{\scalebox{1.5}{$\downarrow$}}
            \vspace{-5pt}
        \end{center}

        {\color{tagpurple}\texttt{<tool\_call>}}
        \hlpurple{\texttt{rg -F "Just Can't Get Enough" corpus/zh | head -n 5}}
        {\color{tagpurple}\texttt{</tool\_call>}} \\

        {\color{taggreen}\texttt{<tool\_response>}} \\
        \textit{[lang=zh]} \\
        \texttt{[ZH]} Doc 1 : \hlgreen{《Just Can't Get Enough》}是英国电子乐队
        \hlorange{流行尖端（Depeche Mode）}
        演唱的歌曲，于1981年发行。 \\[2pt]
        \texttt{[ZH]} : 英国电子乐队\hlorange{流行尖端（Depeche Mode）}的早期代表作包括
        \hlgreen{《Just Can't Get Enough》}
        和《New Life》等歌曲。 \\[2pt]
        \texttt{[ZH]} Doc 3 : 《Speak \& Spell》是\hlorange{流行尖端}的首张录音室专辑，其中收录了
        \hlgreen{《Just Can't Get Enough》}。 \\
        {\color{taggreen}\texttt{</tool\_response>}} \\


        {\color{tagblue}\texttt{<think>}}
        \hlblue{重新检索得到的多个中文文档均表明，
        《Just Can't Get Enough》的演唱者是英国电子乐队 Depeche Mode，
        中文名为“流行尖端”。}
        {\color{tagblue}\texttt{</think>}} \\

        {\color{tagorange}\texttt{<answer>}}
        \hlorange{流行尖端（Depeche Mode）}
        {\color{tagorange}\texttt{</answer>}}

    \end{tcolorbox}

    \caption{Case study of Active-Anchor grounding.
    The initial exact search fails because the queried surface form
    \textit{I Just Can't Get Enough} does not match the corpus-native title.
    Active-Anchor probes the English pivot corpus with the failed lexical form,
    recovers the canonical title \textit{Just Can't Get Enough},
    and reuses it to successfully retrieve the answer from the Chinese collection.
    Green blocks denote useful evidence or recovered entity forms.}
    \label{fig:case_active_anchor}
\end{figure*}
\end{CJK*}
\end{document}